# ALReLU: A different approach on Leaky ReLU activation function to improve Neural Networks Performance

**Stamatis Mastromichalakis[1]**

[1]London South Bank University / IST College,
Pireos 72, GR-18346, Moschato, Athens, Greece
Email: stamatis@tmnetworks.gr

**ABSTRACT**

*Despite the unresolved 'dying ReLU problem', the classical ReLU activation function (AF) has been extensively applied in Deep Neural Networks (DNN), in particular Convolutional Neural Networks (CNN), for image classification. The common gradient issues of ReLU pose challenges in applications on academy and industry sectors. Recent approaches for improvements are in a similar direction by just proposing variations of the AF, such as Leaky ReLU (LReLU), while maintaining the solution within the same unresolved gradient problems. In this paper, the Absolute Leaky ReLU (ALReLU) AF, a variation of LReLU, is proposed, as an alternative method to resolve the common 'dying ReLU' on NN-based algorithms for supervised learning. The experimental results demonstrate that by using the absolute values of LReLU's small negative gradient, has a significant improvement in comparison with LReLU and ReLU, on image classification of diseases such as COVID-19, text and tabular data classification tasks on five different datasets.*

## 1. INTRODUCTION

Although recent developments of AFs for Shallow and Deep Learning Neural Networks (NN), such as the QReLU/m-QReLU (Parisi et al., 2020a) and m-arcsinh (Parisi et al., 2020b), the repeatable and reproducible functions have remained very limited and confined to three activation functions regarded as 'gold standard' (Parisi et al., 2020b). The sigmoid and tanh are well-known for their common vanishing gradient issues and only ReLU function seems to be more accurate and scalable for deep neural networks, despite its 'dying ReLU' problem, which has recently appears to been solved by (Parisi et al., 2020a). Earlier, many other methods are proposed in order to fix these known problems. Some of them are multiple variations of the ReLU AF, such as the Leaky ReLU (LReLU), the Parametric ReLU (PReLU), the Randomised ReLU (RReLU) and the Concatenated ReLU (CReLU). Furthermore, Leaky ReLU (LReLU) introduced (Maas et al. 2013) by providing a small negative gradient for negative inputs into a ReLU function, instead of being 0. A constant variable $\alpha$, with a default value of 0.01, was used to compute the output for negative inputs. Using this modification, LReLU can lead to small improvements in classification performance when compared to the ReLU AF. But the most of the above mentioned AFs are tend to have lack of robustness with classification tasks of varying degrees of complexity, e.g., slow or lack of convergence (Vert and Vert, 2006) (Jacot et al., 2018), caused by trapping at local minima (Parisi et al., 2020b). Amongst the mentioned AFs, only the ReLU is really applicable on NNs, with its novel quantum variations (QReLU and m-QReLU) found more scalable than its traditional version only recently (Parisi et al., 2020a).

In this work, a new approach of LReLU is proposed in order to solve the common gradient vanishing and 'dying ReLU' problems, by using the absolute values of the small negative gradient used on LReLU. This method leads in a significance improvement on training and classification procedures as it is concluded from the results of the numerical evaluation performed. Quantitative evaluation metrics such as accuracy, AUC, recall, precision and F1-score have been computed to reveal the performance of the proposed technique and to provide the necessary criteria for a reliable objective evaluation of the method.

The outline of this paper is as follows: Section 2 contains one of the main contributions of this work, which includes the implementation of ALReLU in Keras. Section 3 presents experimental results of the proposed AF, including an evaluation of the accuracy of the training. Also it is compared to other well defined AFs in the field. Finally, discussion and the main conclusions of the work are devoted on Section 4.

## 2. METHODS AND ALGORITHMS

### 2.1 Datasets used from Kaggle and NN models hyperparameters

The following data sets for image, text and tabular data classification were used in the experiments described and discussed in this study:

- COVID-19 X-Rays and Pneumonia Datasets (Kaggle) having 372 and 5,863 X-Ray images respectively.(https://www.kaggle.com/bachrr/covid-chest-xray?select=images, https://www.kaggle.com/paultimothymooney/chest-xray-pneumonia)
- Parkinson Detection Spiral Drawing & Waves (Kaggle) having 204 total images of Waves and Spirals draws images (https://www.kaggle.com/kmader/parkinsons-drawings)
- Histopathologic Cancer Detection Dataset having 57,458 32x32 colour images and it was used in Kaggle competition on 2019 (https://www.kaggle.com/c/histopathologic-cancer-detection/data)
- Quora Insincere Questions Detection Dataset consist by 1,681,928 unique questions in total labeled with 0 if considered not toxic and 1 if considered toxic. It was also used in Kaggle competition on 2019 (https://www.kaggle.com/c/quora-insincere-questions-classification/data)
- Microsoft Malware Prediction was used in Kaggle competition on 2019 too, and has 16,774,736 records in tabular format (https://www.kaggle.com/c/microsoft-malware-prediction/data)

The CNN-related hyperparameters to train the datasets of COVID-19, Spiral Drawing & Waves and Histopathologic Cancer Detection data set is a deep CNN and has the following layers:

- five convolutional layers, the first of which has kernel size of 5x5 and the four last 3x3
- the following convolutional filters for each of the two convolutional layers respectively (in order from the first layer to the last one): 32, 64, 128, 256, 512
- Max Pooling and Batch Normalization is applied after all convolutional layers;
- Dropout layer also after all convolutional layers leveraged with dropout rates (in order from the first layer to the last one): 0.1, 0.2, 0.3, 0.4, 0.5
- The AFs were also applied after all convolution layers
- A Global Average Pooling Layer followed by AF, Batch Normalization and dropout with rate 0.3
- A Dense layer of size 256 followed by AF, Batch Normalization and dropout with rate 0.4
- a final Dense layer with softmax activation, having as size of neurons the number of output classes of each the dataset

On Quora Insincere Questions Dataset, it was used a Bidirectional LSTM CNN, set are as follows:
- As embedding layer, it was used the mean of both Glove and Paragram embeddings (these embeddings are included in the Quora Kaggle Dataset that is referred before.)
- A Spatial Dropout at 0.2 rate
- A Bidirectional LSTM layer with 128 RNN units,
- four convolutional layers, each of which has a kernel size 1, 2, 3, 5 and filters of 100, 90, 30, 12 respectively
- Each convolution layers is followed by the AF
- four Global Max Pooling layer
- A Dense layer of 200, followed by AF dropout (0.2) and batch normalization
- a final Dense layer with softmax activation, having 2 neurons as for each class (toxic, not toxic)

A shallow neural network model is used to train Microsoft Malware Prediction dataset with the following properties:
- A Dense layer of size 100, followed by dropout rate of 0.4, Batch Normalization and AF
- Another Dense layer of size 100, followed by dropout rate of 0.4, Batch Normalization and AF
- a final Dense layer with softmax activation, having 2 neurons as for each class (has malware detection, not has malware detection)

## 2.2 The ALReLU AF as an LReLU alternative

Rectified Linear Unit, or ReLU, is one of the most common AFs used in NNs today. It is commonly used on between layers to add nonlinearity in order to handle more complex and nonlinear datasets. Fig. 1, demonstrates the ReLU that can be expressed as follows (Eq. (2) is ReLU derivative):

$$f(x) = \begin{cases} 0 \forall \ x < 0 \\ x \ \forall \ x > 0 \end{cases} \quad (1)$$

$$\frac{dy}{dx} f(x) = \begin{cases} 0 \forall \ x < 0 \\ 1 \ \forall \ x > 0 \end{cases} \quad (2)$$

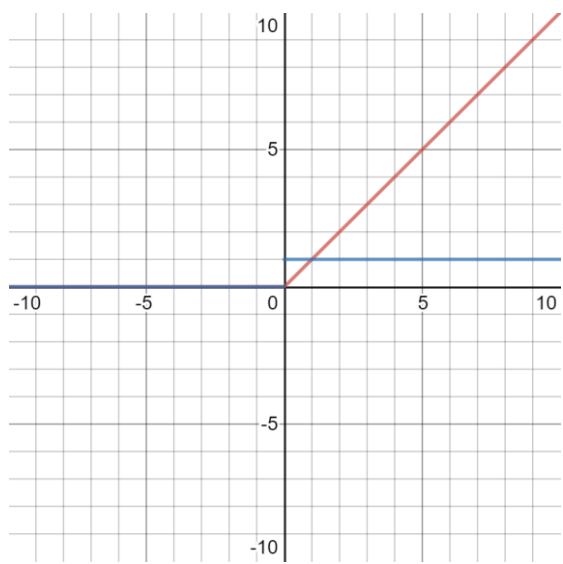

**Figure 1:** Red: ReLU AF, Blue: ReLU Derivative

Despite its success, especially on DNN, ReLU has a few issues. Firstly, ReLU is not continuously differentiable. At x=0, the gradient cannot be computed. Although it is not a serious problem, it slightly impact training performance.

Moreover, ReLU sets all values < 0 to zero. This can benefit on sparse data, but since the gradient of 0 is 0 and hence neurons arriving at large negative values cannot recover from being stuck at 0. The neuron effectively dies and hence the problem is known as the 'dying ReLU' problem. This can leads the network essentially stops learning and underperforms.

Despite appropriate initialization of the weights to small random values, with large weight updates, the summed input to the traditional ReLU AF is always negative, although the input values fed to the NN. Current improvements to the ReLU, such as the LReLU, allow for a more non-linear output to either account for small negative values or facilitate the transition from positive to small negative values, without eliminating the problem though.

The LReLU is trying to solve these problems by providing a small negative gradient for negative inputs into a ReLU function. Fig. 2 and Eqs. (3) and (4) demonstrate the LReLU and its derivative.

$$f(x) = \begin{cases} x \ \forall \ x > 0 \\ ax \ \forall \ x \leq 0 \\ \text{where } \alpha = 0.01 \end{cases} \quad (3)$$

$$\frac{dy}{dx} f(x) = \begin{cases} 0.01 \ \forall \ x < 0 \\ 1 \ \forall \ x > 0 \end{cases} \quad (4)$$

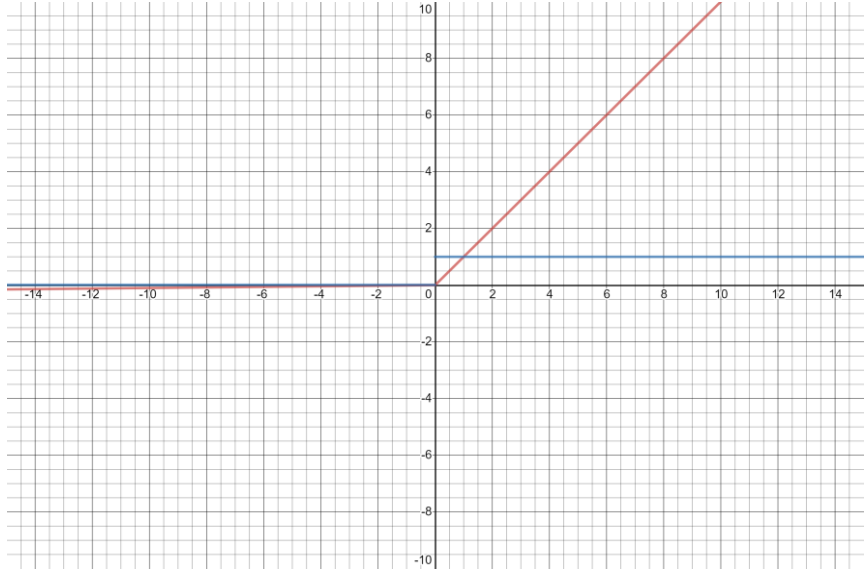

**Figure 2:** Red: LReLU AF, Blue: LReLU Derivative

Although theoretically LReLU is solving the 'dying ReLU', it is not actually proven to improve the classification performance. Indeed, in several studies the LReLU performance is the same or lower with ReLU. Consequently, this study investigates the development of a modified version of ReLU AF to obviate the problem of the 'dying ReLU' by using absolute values of LReLU, i.e., by achieving a positive solution where the solution was negative.

Actually, the idea to use absolute value rectification was first introduced by (Jarrett et al., 2009), and it used for object recognition from images where it makes sense to seek features that are invariant under a polarity reversal of the input illumination (GoodFellow et al., 2016). This work, goes a step further, with the proposal to use the absolute values of LReLU.

Fig. 3 and Eqs. (5) and (6) show the modified version that uses LReLU absolute values (ALReLU) and its derivative:

$$f(x) = \begin{cases} x \ \forall \ x > 0 \\ |ax| \ \forall \ x \leq 0 \\ \text{where } \alpha = 0.01 \end{cases} \quad (5)$$

$$\frac{dy}{dx} f(x) = \begin{cases} -0.01 \ \forall \ x < 0 \\ 1 \ \forall \ x > 0 \end{cases} \quad (6)$$

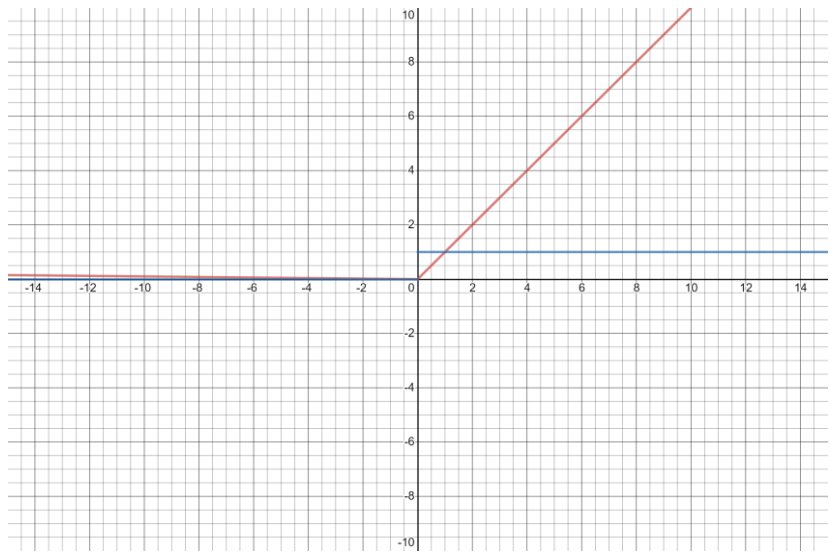

**Figure 3:** Red: ALReLU AF, Blue: ALReLU Derivative

The derivative of ALReLU on x<0 is the negative derivative of LReLU. On x>0 it is has the same derivative.

On the other hand, in Fig. 4 it is demonstrated the QReLU (L. Parisi, at al., 2020a). It can be considered that this AF and its derivatives are a little similar with the proposed one.

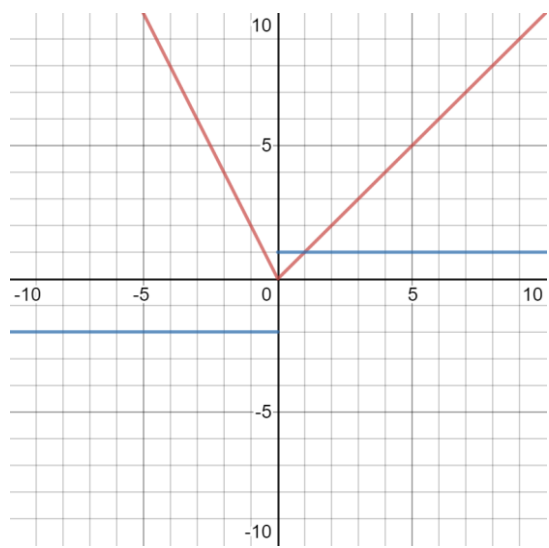

**Figure 4:** Red: QReLU AF, Blue: QReLU Derivative

The main difference is that ALReLU has smaller value and derivative. In a theoretically perspective, the ALReLU has also the properties of QReLU, such as the advantage of superposition and entanglement principles. However, this claim is only in theory and it is not proven in this paper. In fact, this work demonstrates the advantages of the little modification of LReLU that leads to the proposed ALReLU AF. The experiments and results on Section 3, indicate the significance impact of this modification.

The following code snippets, show the Keras implementation of the proposed AF, as well as its usage after the Convolution Layers. The derivatives and gradients of AFs are automatically calculated in TensorFlow 2, so they have not implemented manually.

**Listing 1:** A snippet of code in Python (Keras) with ALReLU implementation and usage
______________________________________________________________________

```python
from tensorflow.keras import backend as K
from tensorflow.keras.layers import Input, Conv2D, Lambda
from tensorflow.keras.utils import get_custom_objects
def ALReLU(x):
    alpha = 0.01
    return K.maximum(K.abs(alpha*x), x)

get_custom_objects().update({'ALReLU':
tf.keras.layers.Activation(ALReLU)})

conv = Conv2D(32, (5, 5))(visible)
conv_act = ALReLU(conv)
conv_act_batch = BatchNormalization()(conv_act)
conv_maxpool = MaxPooling2D()(conv_act_batch)
conv_dropout = Dropout(0.1)(conv_maxpool)
```

**Listing 2:** A snippet of code in Python (Keras) with ALReLU usage in Bidirectional LSTM CNN
______________________________________________________________________

```python
x1 = Conv1D(filters=filters[0], kernel_size=1,
            padding='same')(x)
x1 = Activation(ALReLU)(x1)
x2 = Conv1D(filters=filters[1], kernel_size=2,
            padding='same')(x)
x2 = Activation(ALReLU)(x2)
x3 = Conv1D(filters=filters[2], kernel_size=3,
            padding='same',)(x)
x3 = Activation(ALReLU)(x3)
x4 = Conv1D(filters=filters[3], kernel_size=5,
            padding='same')(x)
x4 = Activation(ALReLU)(x4)

x1 = GlobalMaxPool1D()(x1)
x2 = GlobalMaxPool1D()(x2)
x3 = GlobalMaxPool1D()(x3)
x4 = GlobalMaxPool1D()(x4)

c = Concatenate()([x1, x2, x3, x4])
x = Dense(200)(c)
x = Activation(ALReLU)(x)
x = Dropout(0.2)(x)
x = BatchNormalization()(x)
x = Dense(2, activation="softmax")(x)
```

## 3. EXPERIMENTAL STUDY AND RESULTS

Since the NN models are trained for given datasets, for estimating the performance and the classification accuracy a 9-Fold validation used in COVID-19 dataset and a 5-Fold cross-validation for other datasets. Cross-validation is a statistical method of evaluating and comparing learning algorithms by avoiding overfitting. The K-Fold validation procedure has been executed 4 times for every NN model and dataset. The average of these measures is computed and demonstrated in this section. The results support the theoretical superiority of the proposed ALReLU AF for image and text classification tasks, as well as tabular data classification. The classification performance results are demonstrated in Table 1 and are described above:

| | Performance measures | ALReLU (this study) | ReLU | Leaky ReLU |
|---|---|---|---|---|
| COVID-19 X-Rays (Kaggle) | Weighted Precision | 80% | 64% | 47% |
| | Accuracy | 66.13% | 64.52% | 59.68% |
| | Weighted Recall | 66% | 64% | 60% |
| | AUC | 91.12% | 92.3% | 92.38% |
| | Weighted F1 | 56% | 56% | 50% |
| Spiral Drawings+Waves (Kaggle) | Weighted Precision | 74% | 70% | 61% |
| | Accuracy | 72.73% | 68.18% | 59.09% |
| | Weighted Recall | 73% | 68% | 59% |
| | AUC | 66.94% | 71.9% | 74.38% |
| | Weighted F1 | 73% | 68% | 57% |
| Histopathologic Cancer Detection (Kaggle) | Weighted Precision | 88% | 89% | 89% |
| | Accuracy | 86.69% | 87.34% | 87.48% |
| | Weighted Recall | 87% | 87% | 87% |
| | AUC | 95.3% | 95.45% | 95.21% |
| | Weighted F1 | 87% | 87% | 87% |
| Quora Insincere Questions (Kaggle) | Weighted Precision | 96% | 96% | 96% |
| | Accuracy | 96.38% | 96.34% | 96.36% |
| | Weighted Recall | 96% | 96% | 96% |
| | AUC | 97.33% | 97.26% | 97.26% |
| | Weighted F1 | 96% | 96% | 96% |
| Microsoft Malware Prediction (Kaggle) | Weighted Precision | 65% | 65% | 65% |
| | Accuracy | 64.8% | 64.68% | 64.57% |
| | Weighted Recall | 65% | 65% | 65% |
| | AUC | 70.89% | 70.87% | 70.73% |
| | Weighted F1 | 65% | 65% | 65% |

**Table 1:** Classification performance measures for ALReLU, pure ReLU and LReLU, various datasets

- ***For CNN train and classification on COVID-19 X-Rays classification***: All the evaluation metrics (Weighted Precision, Accuracy, Weighted Recall, AUC, F1) are better than original ReLU and LReLU activation functions. This demonstrated in the Table 1 first row dataset "COVID-19 X-Rays (Kaggle)".

- ***For CNN train and classification on Spiral Drawings and Waves for Parkinson disease prediction***: The results show a better performance of the proposed AF on Weighted Precision, Accuracy, Weighted Recall, F1. Only the AUC was a little lower (66.94% VS 71.9% and 74.38%) comparing with original ReLU and LReLU. (Table 1, second row "Spiral Drawings+Waves (Kaggle)".

- ***For CNN train and classification of Histopathologic Cancer Detection Dataset***: The 3 AF have almost similar performance in this case. The proposed AF has a little lower results on Weighted Precision (88% VS 89% and 89%), Accuracy (86.69% VS 87.34% and 87.48%), and AUC (95.3% VS 95.45% and 95.21%). The other metrics are the same. (Table 1, third row "Histopathologic Cancer Detection (Kaggle)").

- ***For LSTM CNN train and classification of Quora Insincere Questions***: In this text classification task, the 3 AF have also similar performance. The proposed AF has a little better results on Accuracy and AUC. The other metrics are the same. (Table 1, third row "Quora Insincere Questions (Kaggle)").

- ***For NN train and classification of Microsoft Malware Prediction Dataset***: In this tabular data classification task the proposed AF has a little better results on Accuracy and AUC. The other metrics are the same. (Table 1, third row "Microsoft Malware Prediction (Kaggle)").

## 4. CONCLUSION

In this paper, a different approach on LReLU function was demonstrated, by employing the absolute values of its negative gradient. It was proven a more accurate and robust AF for Neural Networks that used in image, text and tabular data classification. The proposed AF is merely centering on the improvement of the classification accuracy in sections wherein high accuracy and reliability is very important to be achieved, such as in the medical sector. Numerous tests were performed in order to validate the theoretical framework developed in this paper. In fact, the proposed method is used in several experiments of training and classification, on different datasets, and it shows superiority to the well-established ReLU and LReLU in terms of ROC/AUC metrics, recall, precision, F1 scores and accuracy. The important conclusion of these results, is the proposal of an alternative method for solving the common dying and vanishing gradient problems, a serious flag on NN training.

# 5. REFERENCES

Luca Parisi, Daniel Neagu, Renfei Ma, and Felician Campean. Qrelu and m-qrelu: Two novel quantum activation functions to aid medical diagnostics. arXiv preprint arXiv:2010.08031, 2020a

Luca Parisi, Renfei Ma, Narrendar RaviChandran, Matteo Lanzillotta: hyper-sinh: An Accurate and Reliable Function from Shallow to Deep Learning in TensorFlow and Keras arXiv preprint arXiv:2011.07661, 2020b

R´egis Vert and Jean-Philippe Vert. Consistency and convergence rates of one-class svms and related algorithms. Journal of Machine Learning Research, 7(May):817–854, 2006.

Arthur Jacot, Franck Gabriel, and Cl´ement Hongler. Neural tangent kernel: Convergence and generalization in neural networks. In Advances in neural information processing systems, pages 8571–8580, 2018.

Maas, A. L., Hannun, A. Y., & Ng, A. Y. (2013). Rectifier Nonlinearities Improve Neural Network Acoustic Models. Proceedings of the 30th International Conference on Machine Learning. Atlanta, Georgia, USA: JMLR: W&CP.

Jarrett, K., Kavukcuoglu, K., Ranzato, M., and LeCun, Y. (2009). What is the best multi-stage architecture for object recognition? In *ICCV'09* . 16, 24, 27, 174, 193, 226, 363, 364, 523

Deep Learning (Ian J. Goodfellow, Yoshua Bengio and Aaron Courville), MIT Press, 2016.